# *MTCurv*: Deep learning for direct microtubule curvature mapping in noisy fluorescence microscopy images

Achraf Ait Laydi[1,2][0009-0004-8898-1903], Sidi Mohamed Sid'El Moctar[2][0009-0009-5490-1916], Yousef El Mourabit[1][0000-0001-7851-3816], and Hélène Bouvrais[2][0000-0003-1128-1322]

[1] TIAD Laboratory, Sciences and Technology Faculty, Sultan Moulay Slimane Univ., Beni Mellal, Morocco.
[2] CNRS, Univ. Rennes, IGDR (Institut de Génétique et Développement de Rennes), UMR6290, 35000 Rennes, France

**Abstract.** Accurate quantification of the geometry of curvilinear biological structures is essential for understanding cellular mechanics and disease-related morphological alterations. Microtubule curvature is a key descriptor of filament rigidity and mechanical perturbations. However, reliable curvature extraction from fluorescence microscopy images remains challenging due to noise, low contrast, and partial filament visibility. Existing approaches rely on segmentation pipelines with pre- or post-processing, which are highly sensitive to segmentation errors and often fail under adverse imaging conditions. In this work, we propose *MTCurv*, a deep learning framework for direct, segmentation-free regression of microtubule curvature maps from noisy microscopy images. Leveraging a synthetic dataset with pixel-wise curvature annotations, we reformulated curvature estimation as a regression problem and adapted an attention-based residual U-Net. To reduce hallucinations and enforce spatial coherence, we introduced a gradient-aware loss combining Mean Squared Error with a gradient consistency term. Beyond model and loss design, we evaluated commonly used regression and image quality metrics, revealing that many perceptual and blind metrics are poorly suited for curvature estimation. Correlation-based metrics, particularly Spearman correlation, emerged as more reliable indicators of curvature prediction quality. Experiments on two datasets of increasing difficulty demonstrated that *MTCurv* accurately recovers local microtubule curvatures, even in the presence of background fluorescence (Spearman = 0.90), while performance degrades for faint microtubule extremities (Spearman = 0.74). Ablation studies highlighted the contribution of both residual encoding and attention-based decoding. Overall, this work provides a practical tool for filament curvature analysis and methodological insights for geometry-aware regression in biomedical imaging. Datasets and code are made available.

## 1 Introduction

Exploring the geometries of cellular structures and organs is essential in biological research, medical diagnostics, and the development of targeted therapies. Curvilinear structures, such as cytoskeletal filaments, nerves, and blood vessels, are central to numerous biological processes, and their shapes and deformations often reflect underlying physiological or pathological conditions. Consequently, accurate characterization of their geometric features in biomedical images is crucial across many application domains. For example, quantifying retinal blood vessel tortuosity supports the diagnosis of several diseases, including retinopathy, hypertension, diabetes, and cardiovascular, renal, or neurological disorders [1, 2]. Likewise, the morphology of neurons and nerves can serve as an informative readout of disease state [3, 4]. In life science research, studying cytoskeletal filaments (microtubules, actin, and intermediate filaments) is fundamental not only for understanding cell function, tissue formation, and organismal development [5, 6], but also for identifying molecular targets and assessing drug efficacy for instance in the context of neurodegenerative diseases and cancers [7, 8]. Overall, accurately extracting geometric descriptors of curvilinear structures provides powerful biomarkers for early diagnosis, monitoring disease progression, and elucidating essential aspects of human biology.

Recent advances in imaging technologies have enabled increasingly detailed observations of biological structures. However, extracting reliable geometric information remains challenging especially under low signal-to-noise conditions or when filaments are only partially visible. Traditional approaches rely on segmenting the structures prior to quantifying their geometric properties. When filaments are clearly visible, these multi-step methods can be effective although time-consuming and often dependent on handcrafted parameters [9-17]. Yet, in many cases, filaments are only partially visible or ambiguously defined, leading to imperfect segmentation [18] and restricting analyses to qualitative observations, even though quantitative measurements would be more informative. We foresaw that the deep learning algorithms accounting for the contextual information within the images to predict features will succeed in estimating geometric properties by considering the neighboring pixels. By avoiding the segmentation step, this original approach may be relevant in images where curvilinear structures are faint.

To challenge this hypothesis, we aimed to extract the longitudinal curvatures of microtubules directly from fluorescence microscopy images. We leveraged a synthetic dataset of fluorescent microtubule networks previously employed for segmentation task [19, 20]. We generated ground-truth curvature maps from these synthetic images, enabling controlled evaluation of our method under varying noise levels. Microtubule curvature in cells is a key quantitative readout for assessing perturbations of microtubule rigidity [10, 14, 15]. A reliable curvature extraction tool would therefore be valuable to a broad community of cell and developmental biologists studying filament mechanics. Our task formulates microtubule curvature estimation as a dense regression problem, which fundamentally differs from most deep learning regression tasks in microscopy that focus on image restoration, denoising, or super-resolution [21-23]. Instead of predicting visually plausible images, the objective here is to recover a geometrically meaningful physical quantity that is highly sensitive to local spatial variations. To address this challenge, we adapted an attention-based residual U-Net architecture [20], and we refer to the resulting framework as *MTCurv*.

This work also highlights the methodological challenges associated with training and evaluating geometry-aware regression models. We show that commonly used pixel-wise regression or perceptual metrics are often inadequate for assessing curvature prediction quality, and we systematically analyzed alternative evaluation strategies that better align with the underlying geometry. To further improve spatial coherence and reduce spurious predictions, we introduce a gradient-aware loss that complements the Mean Squared Error by enforcing consistency in curvature variations [24, 25]. The contributions of this paper are:

- Introducing *MTCurv*, a segmentation-free deep learning framework for direct microtubule curvature regression from noisy fluorescence microscopy images.
- A gradient-aware loss function enforcing spatial consistency in curvature prediction.
- A critical evaluation of commonly used regression and image quality metrics, identifying correlation-based measures as more reliable curvature-quality indicators.
- An ablation study to quantify the impact of residual encoding and attention-based decoding on curvature estimation accuracy.

# 2 Method

## 2.1 Datasets

We previously developed two synthetic datasets, *MicSim_FluoMT*, which are composed of microscopy images displaying fluorescent microtubules and their corresponding masks featuring the microtubule positions [19]. The fluorescence microscopy images simulate astral microtubule networks in the *Caenorhabditis elegans* zygote. These two datasets, called *MicSim_FluoMT* simple and complex, differ in the difficulty of visualizing the extremities of the astral microtubules, to challenge model performance in adverse conditions (Fig. 2A, D). We complemented these datasets by computing the local curvatures along each microtubule (Fig. 2B, E). We used the three-point approach to calculate these curvatures [11]. Curvatures were encoded pixelwise as intensities, which generated curvature maps of the microscopy images. This new dataset, named *MicSim_FluoCurv*, is composed of 1000 image pairs, classified as simple (Fig. 2A-B) or complex (Fig. 2D-E), and is available on Zenodo (doi: 10.5281/zenodo.17990170).

## 2.2 The *MTCurv* architecture

Recently, U-Net-based models have been widely used for segmentation tasks in biomedical images, even for segmenting curvilinear structures (e.g., [20, 26, 27]). For regression problems in biological images (e.g. image restoration or super-resolution), U-Net variants are also employed [21, 22]. We previously proposed a model for segmenting microtubules in noisy images [20], which includes Residual blocks in the U-Net encoder and ASE (Adaptive Squeeze-and-Excitation) attention mechanisms in the U-Net decoder. This approach accounts for image noise variability. Given its strong ability to detect fine structures in noisy backgrounds, we extended the model to a regression setting by adding a linear projection as the final layer. The backbone was optimized with four down-sampling blocks (starting at 32 filters and doubling at each stage), followed by a bottleneck layer with 512 filters, and four corresponding up-sampling blocks. We refer to this architecture as *MTCurv* (Fig. 1), which contains approximately 8 million parameters and is available on Zenodo (doi: 10.5281/zenodo.18156082).

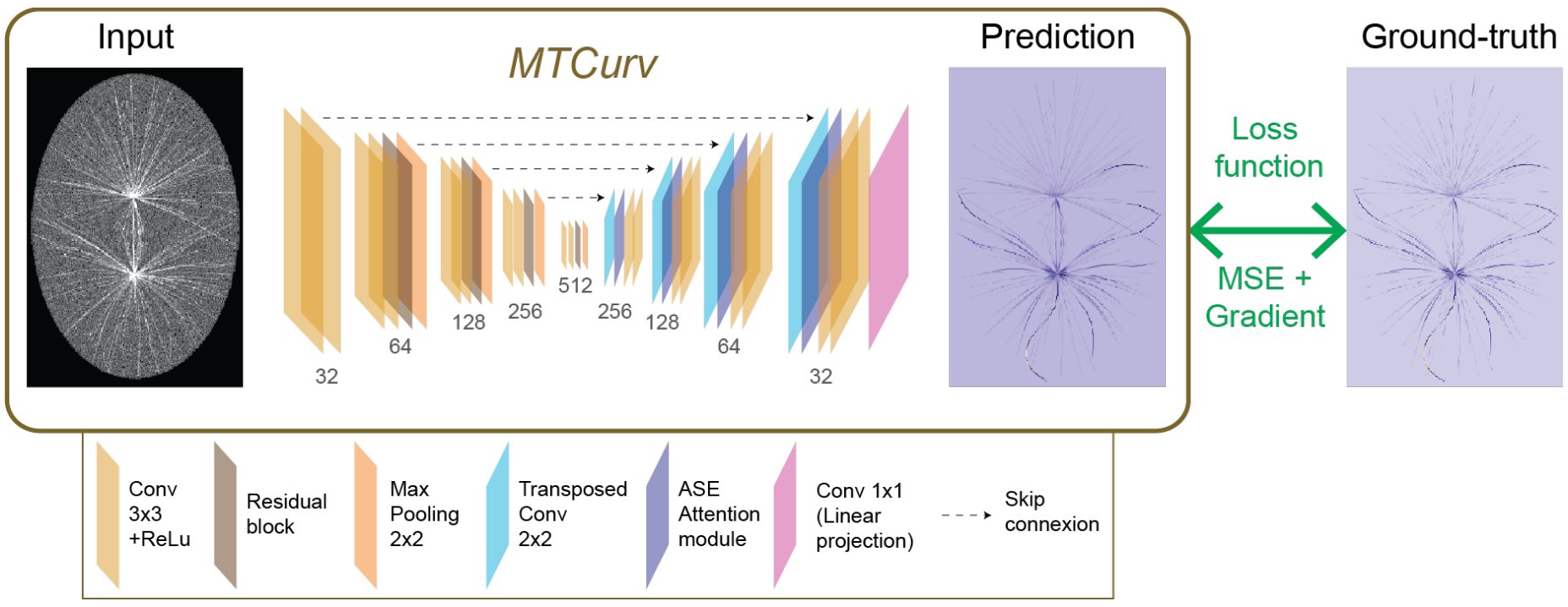

**Fig. 1.** *MTCurv*, a deep-learning framework that extracts directly microtubule curvatures from fluorescence microscopy images using attention-based residual U-Net and MSE-Gradient loss.

### 2.3 Training

We used the Adam optimizer for training, which has proven effective for deep learning applications in biomedical image analysis. The initial learning rate was set to $1\times10^{-3}$. Models were trained for a maximum of 300 epochs with early stopping if the validation loss did not improve for 50 consecutive epochs. The batch size was set to 2, representing a trade-off between GPU memory constraints and training stability. To improve model generalization and robustness to variability in filament orientation and noise patterns, on-the-fly data augmentation was applied during training. Augmentations included random rotations, horizontal and vertical flips, and intensity scaling. These transformations preserve the underlying curvature information while increasing the diversity of training samples. Hyperparameters were selected empirically based on validation performance and kept fixed across all experiments to ensure consistency. For baseline experiments, Mean Squared Error (MSE) loss (Eq. 1) was used, as a standard choice for regression tasks [28]. We then evaluated several composite loss functions, including combinations of MSE with gradient-based and structural terms (cf. Results).

$$MSE = \frac{1}{n}\sum_{i=1}^{n}(y_i - \hat{y}_i)^2 \quad (1)$$

with $y_i$ the true value and $\hat{y}_i$ the prediction.

For dataset splitting, we adopted a random split strategy to ensure fair and reproducible comparisons on the *MicSim_FluoCurv* datasets. Specifically, 80% of the data were used for training (800 images), 10% for validation (100 images), and 10% for testing (100 images). Images from the simple and complex datasets were handled independently. The test set was kept strictly unseen during training and hyperparameter tuning. Model performance was evaluated on the test set by computing the mean and standard deviation (std) of each metric. To further assess robustness and reduce potential bias introduced by a single random split, we performed a 5-fold cross-validation (CV) on the training and validation sets for selected configurations. Performance metrics were averaged across folds, and the corresponding standard deviations were reported. Cross-validation is particularly important with heterogeneous image content, as it provides a more reliable estimate of model generalization.

### 2.4 Assessment metrics and statistics

Since measuring filament geometric features from images is not a standard regression task, it is necessary to identify metrics that can reliably assess model performance [29, 30]. In the context of regression, evaluation metrics can be grouped into five main classes of Image Quality Assessment (IQA): (i) pixel-wise error metrics that measure differences between prediction and ground-truth on a pixel-by-pixel basis, thus focusing on numerical fidelity (e.g., mean absolute error (MAE), mean squared error (MSE), peak signal-to-noise ratio (PSNR)); (ii) statistical metrics that quantify the relationship between prediction and ground truth (e.g., Pearson correlation, Spearman correlation, coefficient of determination ($R^2$), Explained Variance Score (EVS)); (iii) perceptual and structural similarity metrics, which aim to reflect human visual perception and typically show strong correlation with subjective image quality (SSIM, MS-SSIM, GMSD, LPIPS, DISTS, Visual Saliency-based Index (VSI), Visual Information Fidelity (VIF)); (iv) feature-based or distribution-based similarity metrics, which compare higher-level feature embeddings or probability distributions rather than raw pixel values (e.g., Cosine similarity); and (v) no-reference IQA metrics that estimate image quality without requiring a ground-truth reference (e.g., PIQUE, NRQM). We foresaw that some of these metrics' classes may not be appropriate for the curvature estimation, for instance the perceptual and structural similarity metrics or the no reference IQA metrics. For each metric, two-tailed Student's *t*-test with Welch–Satterthwaite correction for unequal variance was performed to assess significant differences between models.

## 3 Results and Discussion

### 3.1 Predictions of microtubule curvature on the *MicSim_FluoCurv* datasets

We tested whether we would succeed in extracting microtubule curvatures without a segmentation step. Since direct extraction of geometric features from microtubules is a regression task, we first used the classical MSE as loss function during the training [28]. We trained *MTCurv* on the *MicSim_FluoCurv* simple dataset and we found that this model succeeds in predicting the whole microtubule curvature map on test images, however with some local curvatures that were imprecise (Fig. 2A-C). Next, we chal-lenged whether we could estimate microtubule curvatures in images from *MicSim_FluoCurv* complex dataset, in which microtubule extremities are faint, and thus more difficult to characterize. When trained on the *MicSim_FluoCurv* complex dataset, *MTCurv* with MSE as loss function partially predicted the microtubule curvature maps of the test images but failed to accurately extract curvatures at microtubule extremities (Fig. 2D-F). In addition, we observed unexpected predictions, hereafter referred to as hallucinations (Fig. 2F, red arrows).

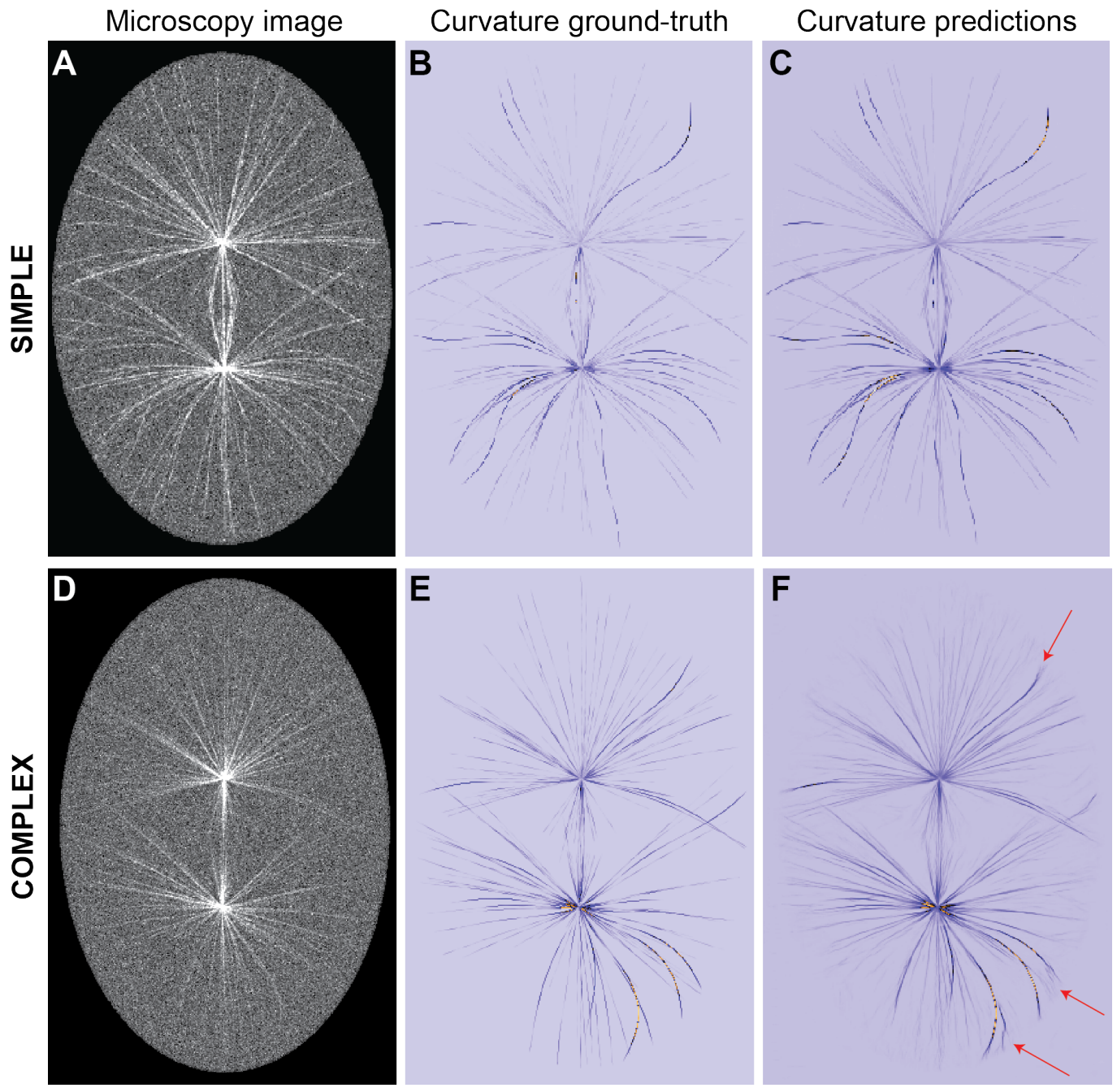

**Fig. 2.** Microtubule curvature extraction from *MicSim_FluoCurv* (**A-C**) simple and (**D-F**) complex datasets using MSE: (A, D) Test microscopy images, (B, E) Ground-truth, and (C, F) Predictions. Curvature maps are visualized using ICA LUT. Red arrows highlight hallucinations.

To better visualize prediction errors, we computed several error maps relevant for regression tasks: (i) the differential error (DE) map (Eq. 2), which highlights regions where curvature is overestimated (bluish colours) or underestimated (reddish colours), (ii) the root squared error (RSE) map, which measures the local quadratic error between the ground truth and the prediction (e.g., [31]), and (iii) the curvature error (CE) map (Eq. 3), which computes the absolute difference between the Laplacians of the ground truth and predicted maps. The latter might be more suited to evaluating local curvature reconstruction.

$$DE(i,j) = T(i,j) - P(i,j) \tag{2}$$

with $T$ the ground truth and $P$ the prediction.

$$CE(i,j) = |\, \Delta P(i,j) - \Delta T(i,j) \,| \quad (3)$$

with $\Delta$ denoting the Laplacian operator.

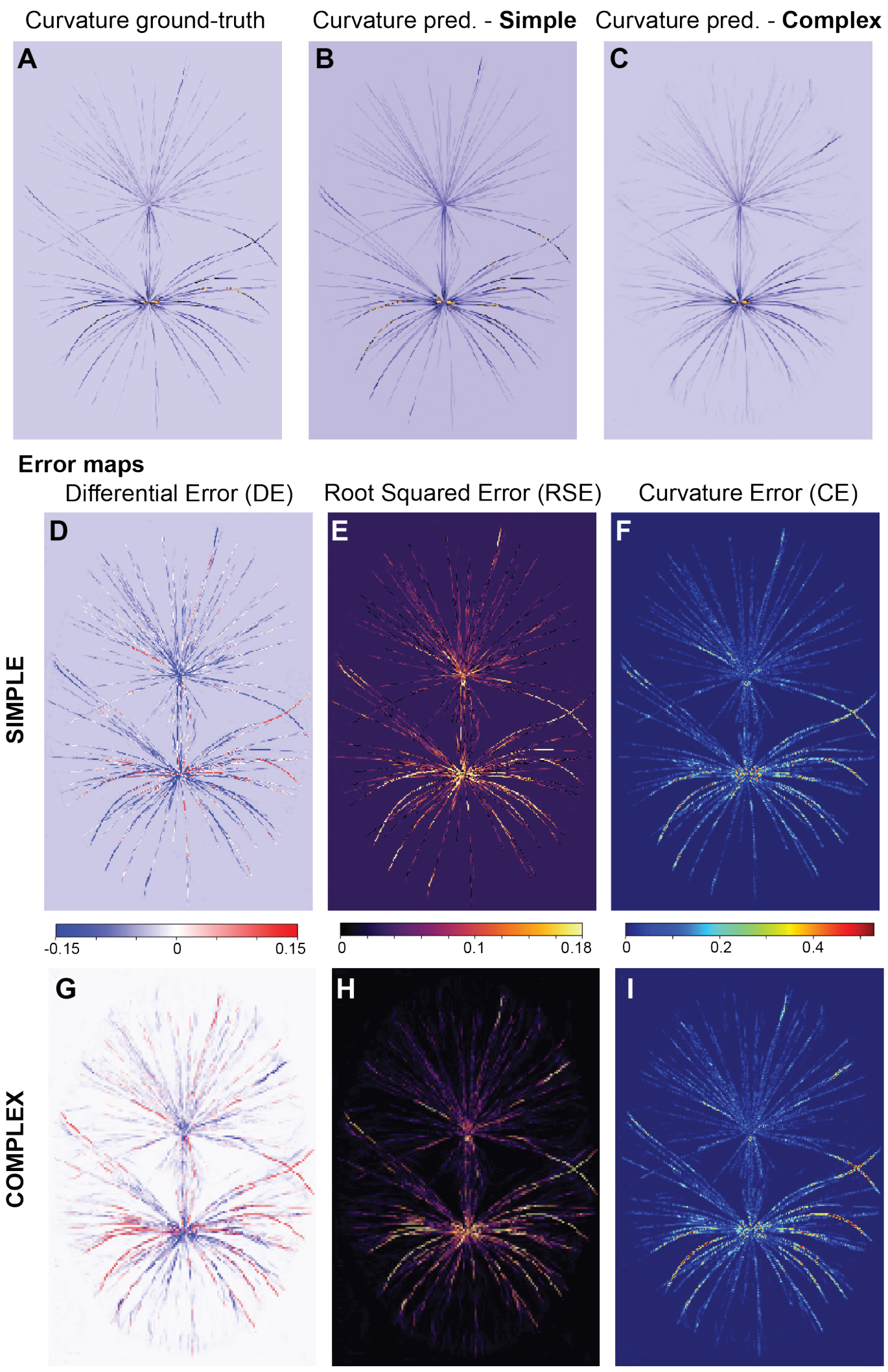

**Fig. 3.** Curvature predictions and error maps using MSE loss. (**A**) Ground truth; (**B, C**) Predictions on the *MicSim_FluoCurv* simple and complex datasets; and Error maps for the (**D-F**) simple and (**G-I**) complex predictions: Differential error, Root Squared Error, and Curvature Error.

We compared the error maps obtained for predictions on the simple and complex datasets to identify the most discriminative representations for quality assessment (Fig. 3). The DE maps were effective at highlighting hallucinations (shown in blue), and at coarsely identifying regions of missed curvature, albeit without fine localization (Fig.

3D, G). In contrast, the CE maps precisely localized microtubule segments with inaccurate curvature estimates, with colour encoding reflecting the error magnitude (Fig. 3F, I). The RSE map did not provide additional informative insights (Fig. 3E, H). These observations suggested that DE and CE maps are complementary tools for visual quality control of curvature estimation. Overall, we concluded that there remained room for improvement in curvature extraction, particularly for the complex dataset. One potential direction was the incorporation of more advanced loss functions.

### 3.2 Assessing appropriate metrics for microtubule curvature prediction.

**Table 1.** *MTCurv* performance trained using the MSE loss on the *MicSim_FluoCurv* simple and complex datasets (mean ± std; Student's *t*-test between simple and complex models).

| Metric category | Metrics | Simple dataset | Complex dataset | *p*-value |
|---|---|---|---|---|
| (i) Pixel-wise | RMSE ↓ | 0.0231 ± 0.0046 | 0.0295 ± 0.0054 | $1.2 \times 10^{-16}$ |
| | NRMSE ↓ [0 1] | 0.6770 ± 0.3723 | 0.8391 ± 0.3129 | 0.001 |
| | MAE | 0.0053 ± 0.0010 | 0.0080 ± 0.0011 | $6.1 \times 10^{-43}$ |
| | SMAPE ↓ [0 200] | 185.20 ± 1.66 | 186.78 ± 1.41 | $1.2 \times 10^{-11}$ |
| | PSNR ↑ [20 50] | 32.9037 ± 1. 7257 | 30.7504 ± 1.6976 | $3.7 \times 10^{-16}$ |
| (ii) Correlation & Statistical | $R^2$ ↑ [0 1] | 0.4044 ± 1.0562 | 0.1990 ± 0.9006 | 0.14 |
| | Pearson ↑ [0 1] | 0.8332 ± 0.0237 | 0.6677 ± 0.0474 | $2 \times 10^{-66}$ |
| | Spearman ↑ [-1 1] | 0.5445 ± 0.0197 | 0.4931 ± 0.0220 | $1.3 \times 10^{-41}$ |
| | EVS ↑ [0 1] | 0.4212 ± 1.0014 | 0.2226 ± 0.8172 | 0.13 |
| (iii) Perceptual & Structural | VSI ↑ [0 1] | 0.9579 ± 0.0093 | 0.9376 ± 0.0105 | $1.1 \times 10^{-32}$ |
| | MS-SSIM ↑ [0 1] | 0.9513 ± 0.0166 | 0.9192 ± 0.0179 | $9 \times 10^{-29}$ |
| | DISTS ↓ | 0.1081 ± 0.0221 | 0.1668 ± 0.0177 | $1.1 \times 10^{-50}$ |
| | VIF ↑ (best = 1) | 0.6820 ± 1.1680 | 0.3690 ± 0.6693 | 0.021 |
| | GMSD ↓ | 0.0729 ± 0.0144 | 0.1077 ± 0.0164 | $3.7 \times 10^{-37}$ |
| (iv) Feature-based simil. | Cosine sim. ↑ [-1 1] | 0.8412 ± 0.0226 | 0.6856 ± 0.0454 | $3.0 \times 10^{-65}$ |
| (v) Blind | PIQE ↓ [0 100] | 75.6938 ± 1.0425 | 69.6480 ± 1.9246 | $3.2 \times 10^{-61}$ |
| | NRQM ↑ [-1 10] | 4.7431 ± 0.7243 | 3.8044 ± 0.5905 | $2.6 \times 10^{-19}$ |

Before testing various loss functions, we aimed to identify metrics that provide reliable readouts of model performance for microtubule curvature mapping, in order to complement visual quality control. Since clear visual differences in prediction quality were observed between the simple and complex datasets, we expected appropriate evaluation metrics to reflect better performance for the model trained on the simple dataset, and sub-optimal performance for the model trained on the complex dataset. We evaluated multiple metrics from the five main classes of IQA (Table 1, §2.4). Pixel-wise metrics revealed significant differences between predictions on the simple and complex da-tasets, with degraded performances on the complex dataset, in agreement with the visual error maps. However, SMAPE values for the simple test images were very high, close to the maximal value of 200%, and did not reflect the correct prediction performance of the simple images compared to the complex ones. Among the correlation and statistical metrics, all four metrics showed degraded performances on the complex dataset compared to the simple one, suggesting that they were reliable metrics. However, $R^2$ and EVS exhibited large standard deviations due to 8 out of 100 test images displaying unexpected values (below 0). These unexpected measurements were associated with images containing fewer curved microtubules. Notably, Pearson correlation values were consistent with the error maps, highlighting overall good curvature predictions on the simple dataset and partially accurate predictions on the complex dataset. The Spearman correlation appeared more sensitive to the imprecise curvatures observed in the simple dataset. Most perceptual and structural metrics were inappropriate, since they yielded values close to optimal for the model trained on the complex dataset, failing to reflect prediction errors. VIF behaved differently by capturing imperfect predictions, albeit with a high standard deviation, similarly to $R^2$ and EVS. Cosine similarity based on features emerged as a promising metric for this task. Blind metrics were difficult to interpret, especially PIQE, which yielded higher values for the simple dataset than for the complex dataset, contrary to expectations. Overall, we retained NRMSE among pixel-wise metrics, along with the correlation and statistical metrics and the Cosine similarity, since they were most consistent with the visual quality control.

### 3.3 Optimizing the loss function

**Table 2.** *MTCurv* performance trained using the MSE-Gradient loss on the *MicSim_FluoCurv* datasets (*t*-tests: (4th col.) simple vs. complex; ($p_{loss}$) MSE vs. MSE-Gradient).

| Metric category | Metrics | Simple dataset | Complex dataset | *p*-value |
|---|---|---|---|---|
| (i) Pixel-wise | NRMSE ↓ | 0.6810 ± 0.3382 | 0.8284 ± 0.1842 | 0.00019 |
| | | $p_{loss}$ = 0.94 | $p_{loss}$ = 0.77 | |
| (ii) Correlation | $R^2$ ↑ | 0.4231 ± 0.9072 | 0.2802 ± 0.4523 | 0.16 |
| | | $p_{loss}$ = 0.89 | $p_{loss}$ = 0.42 | |
| & Statistical | Pearson ↑ | 0.8277 ± 0.0250 | 0.6653 ± 0.0520 | $5.1 \times 10^{-60}$ |
| | | $p_{loss}$ = 0.11 | $p_{loss}$ = 0.73 | |
| | Spearman ↑ | 0.8969 ± 0.0124 | 0.7392 ± 0.0203 | $1.9 \times 10^{-120}$ |
| | | $p_{loss} = 9.5 \times 10^{-181}$ | $p_{loss} = 2.8 \times 10^{-154}$ | |
| | EVS ↑ | 0.4426 ± 0.8385 | 0.3007 ± 0.3822 | 0.13 |
| | | $p_{loss}$ = 0.87 | $p_{loss}$ = 0.39 | |
| (iv) Feature- | Cosine simil. ↑ | 0.8361 ± 0.0239 | 0.6809 ± 0.0492 | $1.1 \times 10^{-60}$ |
| based simil. | | $p_{loss}$ = 0.12 | $p_{loss}$ = 0.46 | |

We hypothesized that a multi-component loss function would improve the curvature mapping, since such approaches have previously been shown to be beneficial in micro-scopic super-resolution or image restoration tasks [22, 23, 32]. Inspired by prior gradient-based losses [24, 25, 33], we defined a simple but effective loss that combines pixel-wise MSE with direct gradient matching between prediction and ground truth to enforce first-order structural consistency. The MSE-Gradient loss reduced the hallucinations observed when using MSE alone, especially for the model trained on the *MicSim_FluoCurv* complex dataset (Fig. 4). The resulting curvature predictions also appeared smoother and more consistent.

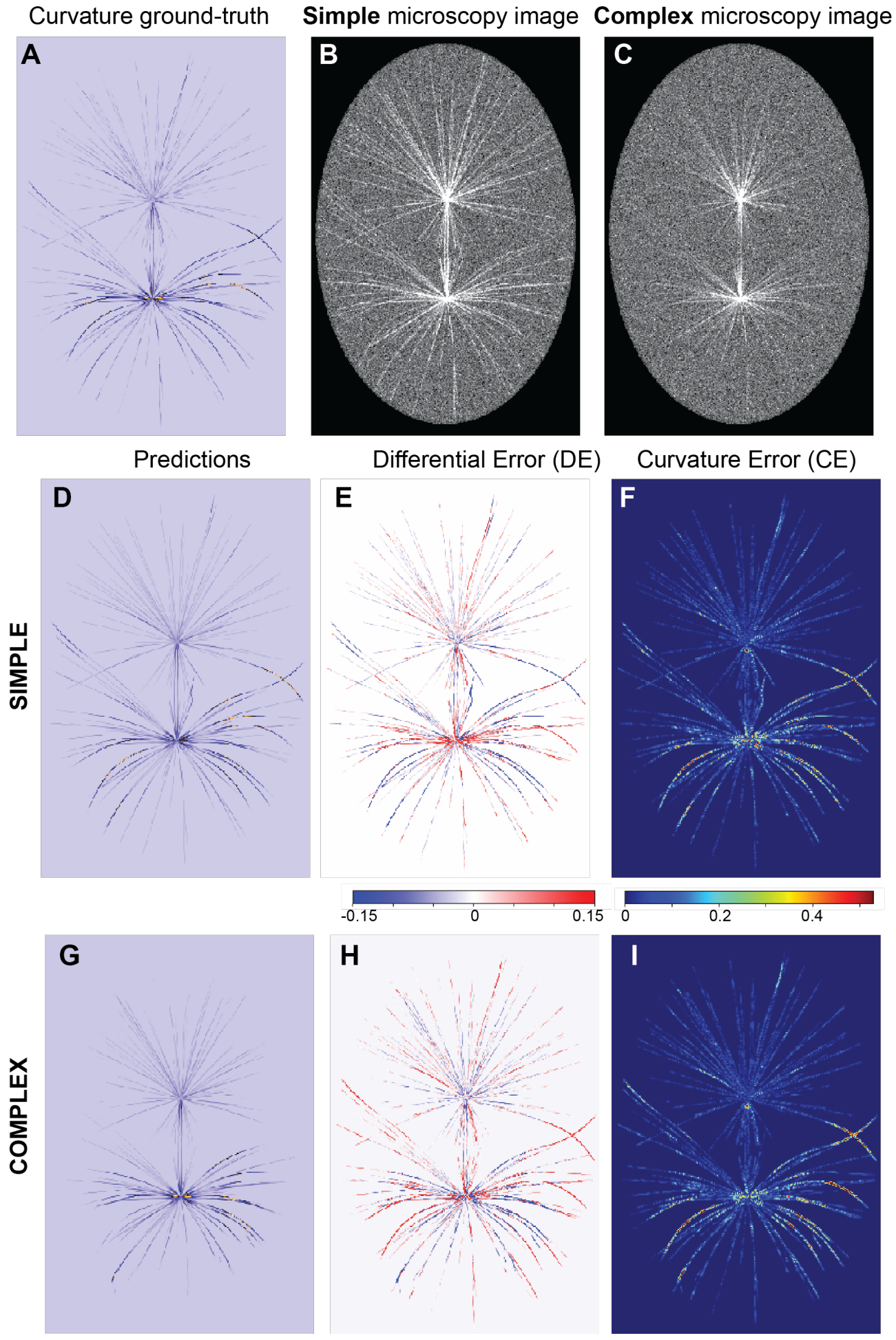

**Fig. 4.** Curvature predictions using the MSE-Gradient loss. (**A**) Ground truth; (**B, C**) *MicSim_FluoCurv* simple and complex images; Results on (**D-F**) simple and (**G-I**) complex images: prediction, differential error, and curvature error.

Model performances, measured using the metrics retained above, were computed for the combined MSE-Gradient loss and confirmed the visual analysis (Table 2). In particular, the Spearman correlation was significantly improved when using the MSE-Gradient loss (Table 2) compared to the MSE loss (Table 1), for models trained either on the simple dataset or on the complex dataset. $R^2$ and EVS scores were also improved when using the MSE-Gradient loss for both models, while NRMSE and Cosine similarity showed improvements only for the model trained on the complex dataset. In contrast, the Pearson correlation was not improved by the use of the MSE-Gradient loss for either model. These results suggested that the Spearman correlation metric is the most sensitive metric for evaluating microtubule curvature predictions. We further tested other combinations of loss functions, including some losses commonly used in image regression tasks such as the MS-SSIM loss, the Huber loss, and a Laplacian consistency term (Table 3). None of these combinations outperformed the MSE-Gradient loss in terms of curvature prediction quality.

**Table 3.** *MTCurv* performances on the *MicSim_FluoCurv* complex dataset with different loss functions (mean ± std of the Spearman correlation on the 100 test images).

| Loss function | MSE+ Grad. | Huber+ Grad. | MSE+ MS-SSIM | MSE+ Laplac. | MSE+ Grad.+ MS-SSIM | Huber+ Grad.+ MS-SSIM | MSE+ Lapl.+ MS-SSIM |
|---|---|---|---|---|---|---|---|
| Spearman | 0.74 ± 0.02 | 0.71 ± 0.02 | 0.48 ± 0.02 | 0.72 ± 0.02 | 0.60 ± 0.03 | 0.60 ± 0.03 | 0.61 ± 0.02 |

Finally, we assessed the robustness and generalization of *MTCurv* by performing a 5-fold cross-validation on the *MicSim_FluoCurv* simple and complex datasets (Table 4). The cross-validation performances were similar to the ones obtained with the random-split approach (Tables 2 & 4). We concluded that the *MTCurv* architecture with the MSE-Gradient loss is a robust approach for extracting microtubule curvatures in microscopy images.

**Table 4.** Cross-validation of *MTCurv* trained using the MSE-Gradient loss on the *MicSim_FluoCurv* simple and complex datasets (mean ± std across the folds).

| Dataset | NRMSE ↓ | $R^2$ ↑ | Pearson ↑ | Spearman ↑ | Cosine sim. ↑ |
|---|---|---|---|---|---|
| Simple | 0.677 ± 0.026 | 0.454 ± 0.083 | 0.827 ± 0.004 | 0.889 ± 0.005 | 0.836 ± 0.004 |
| Compl. | 0.825 ± 0.012 | 0.291 ± 0.032 | 0.652 ± 0.006 | 0.732 ± 0.003 | 0.669 ± 0.006 |

### 3.4 Ablation study

We aimed to evaluate how the architectural components of *MTCurv* contribute to microtubule curvature mapping. We thus compared *MTCurv* with three variant architectures: the baseline U-Net; *MTCurv_noAtt* that mirrors the encoder path by accounting only for the residual blocks; and *MTCurv_noRes* that mimics the decoder path with the inclusion of the ASE attention mechanisms. We performed this ablation study on the *MicSim_FluoCurv* complex dataset using the MSE-Gradient loss. Spearman correlation scores revealed that *MTCurv* outperforms all three variant architectures (Table 5). Among the remaining five reported metrics, four highlighted *MTCurv* as the best-performing architecture, with statistically significant differences observed for the Pearson correlation and the Cosine similarity (Table 5). Visually, the combination of encoding residual blocks and decoding ASE attention mechanisms improved curvature mapping at faint microtubule extremities (Fig. 5). These results demonstrated that enhanced feature extraction and noise handling are crucial for improving microtubule curvature mapping, particularly in highly degraded regions.

**Table 5.** Ablation study using the MSE-Gradient loss on the *MicSim_FluoCurv* complex dataset (mean ± std; Student's *t*-test between *MTCurv* and its variants).

| Metrics | *U-Net* | *MTCurv_noAtt* | *MTCurv_noRes* | *MTCurv* |
|---|---|---|---|---|
| NRMSE ↓ | 0.8380 ± 0.1271 | 0.8418 ± 0.1763 | 0.8309 ± 0.1922 | **0.8284 ± 0.1842** |
| | $p = 0.67$ | $p = 0.63$ | $p = 0.97$ | |
| $R^2$ ↑ | **0.2818 ± 0.2654** | 0.2607 ± 0.4239 | 0.2730 ± 0.4963 | 0.2802 ± 0.4523 |
| | $p = 0.98$ | $p = 0.75$ | $p = 0.91$ | |
| Pearson ↑ | 0.6372 ± 0.0509 | 0.6347 ± 0.0484 | 0.6525 ± 0.0529 | **0.6653 ± 0.0520** |
| | $p = 1.5 \times 10^{-3}$ | $p = 2.6 \times 10^{-5}$ | $p = 0.087$ | |
| Spearman ↑ | 0.3469 ± 0.0255 | 0.4217 ± 0.0239 | 0.4233 ± 0.0209 | **0.7392 ± 0.0203** |
| | $p = 2.7 \times 10^{-180}$ | $p = 2.5 \times 10^{-169}$ | $p = 4.6 \times 10^{-178}$ | |
| EVS ↑ | 0.2957 ± 0.2319 | 0.2759 ± 0.3787 | 0.2905 ± 0.4387 | **0.3007 ± 0.3822** |
| | $p = 0.91$ | $p = 0.65$ | $p = 0.86$ | |
| Cosine sim. ↑ | 0.6575 ± 0.0483 | 0.6552 ± 0.0462 | 0.6712 ± 0.0500 | **0.6809 ± 0.0492** |
| | $p = 8.4 \times 10^{-3}$ | $p = 1.8 \times 10^{-3}$ | $p = 0.17$ | |

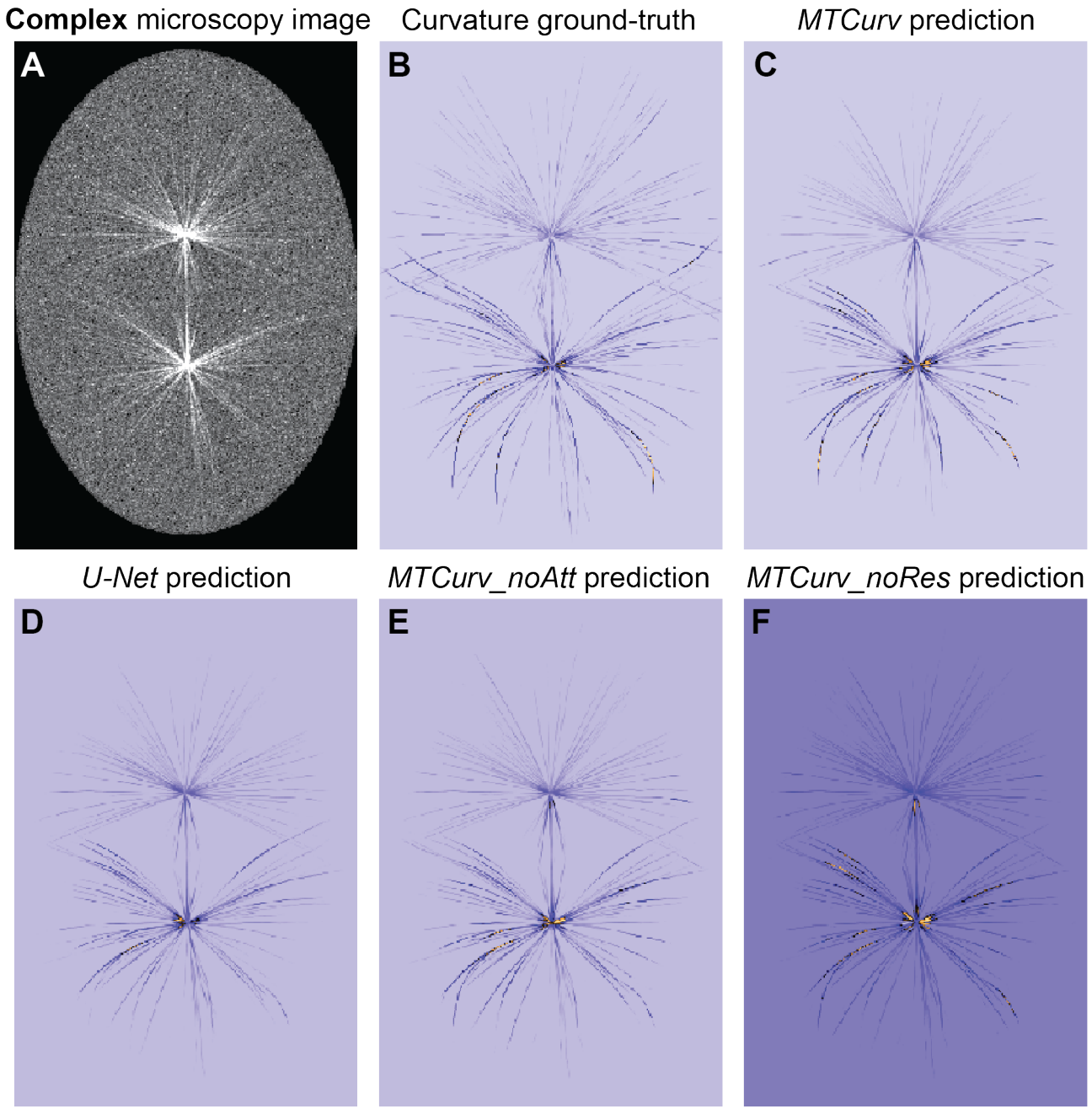

**Fig. 5.** Ablation study performed on the *MicSim_FluoCurv* complex dataset.

## 4 Conclusion

We introduce *MTCurv*, a tool that directly maps microtubule curvatures in microscopy images. This deep-learning framework combines a noise-adaptive-attention-based residual U-Net architecture with a composite loss consisting of a MSE function and a gradient consistency term, and it identifies reliable metrics (notably the Spearman correlation) as well as two error maps (differential error and curvature error) to faithfully assess prediction quality. To our knowledge, this work is among the first to leverage the power of deep learning to directly extract geometric features from biomedical images without requiring a segmentation step. By introducing two noisy image datasets (*MicSim_FluoCurv*), one of which presents increasing levels of difficulty due to barely visible microtubule extremities, we challenged *MTCurv* under adverse conditions that often prevent accurate filament segmentation. Our results highlight the benefits of combining encoding residual blocks with decoding ASE attention mechanisms for extracting microtubule curvatures in low signal-to-noise ratio regions, while keeping the number of model parameters relatively low (i.e., 8 million), compared to recent approaches such as transformers. To further improve the performance of *MTCurv* tool, future work will investigate whether incorporating a statistical-based loss or a cosine similarity as additional loss function components [34] can improve microtubule curvature mapping, since these metrics were found to be reliable indicators of curvature accuracy. We also plan to evaluate *MTCurv* on a broader range of curvature accuracy. We also plan to evaluate *MTCurv* on a broader range of applications to further demonstrate its generalization capabilities. This will require the creation of appropriate training datasets, for example by computing curvature maps from existing binary mask datasets, such as *MicReal_FluoMT* that contains real microscopy images of fluorescent microtubules [35], as well as DRIVE [36] and CORN-1 [37] that include biomedical images of retinal blood vessels and of corneal nerves, respectively. We foresee that *MTCurv* will be relevant for biomedical research, offering new opportunities to investigate cellular mechanics, develop innovative therapeutic strategies, and enhance diagnostic capabilities through more precise characterization of curvilinear biological structures.

**Acknowledgments.** This study was funded by PHC Toubkal 2024 (n° 49945RE), the Agence Nationale de la Recherche (JCJC project MICENN, ANR-22-CE45-001601) and the University of Rennes (Soutien aux collaborations internationales, 2024). The computation servers were funded by the Brittany region (AAP PME 2018-2019 - Roboscope) and by the Agence Nationale de la Recherche (PRCE project SAMIC, ANR-19-CE45-0011). This project was provided with computing AI and storage resources by GENCI at IDRIS thanks to the grants 2024-AD010315542 and 2025-AD010315542R1 on the supercomputer Jean Zay'sV100 partition.

### References

1. Xue, C.C., Li, C., Hu, J.F., Wei, C.C., Wang, H., Ahemaitijiang, K., Zhang, Q., Chen, D.N., Zhang, C., Li, F., Zhang, J., Jonas, J.B., Wang, Y.X.: Retinal vessel caliber and tortuosity and prediction of 5-year incidence of hypertension. Journal of Hypertension 41, 830-837 (2023)

2. Bullitt, E., Zeng, D., Gerig, G., Aylward, S., Joshi, S., Smith, J.K., Lin, W., Ewend, M.G.: Vessel tortuosity and brain tumor malignancy: a blinded study1. Academic radiology 12, 1232-1240 (2005)
3. Cruzat, A., Qazi, Y., Hamrah, P.: In Vivo Confocal Microscopy of Corneal Nerves in Health and Disease. The Ocular Surface 15, 15-47 (2017)
4. Grace, E., Rabiner, C., Busciglio, J.: Characterization of neuronal dystrophy induced by fibrillar amyloid β: implications for Alzheimer's disease. Neuroscience 114, 265-273 (2002)
5. Röper, K.: Microtubules enter centre stage for morphogenesis. Philosophical Transactions of the Royal Society B 375, 20190557 (2020)
6. Matis, M.: The Mechanical Role of Microtubules in Tissue Remodeling. BioEssays 42, 1900244 (2020)
7. Brunden, K.R., Trojanowski, J.Q., Smith III, A.B., Lee, V.M.-Y., Ballatore, C.: Microtubule-stabilizing agents as potential therapeutics for neurodegenerative disease. Bioorganic & medicinal chemistry 22, 5040-5049 (2014)
8. Lafanechère, L.: The microtubule cytoskeleton: An old validated target for novel therapeutic drugs. Frontiers in Pharmacology 13, (2022)
9. Zhang, Z., Nishimura, Y., Kanchanawong, P.: Extracting microtubule networks from superresolution single-molecule localization microscopy data. MBoC 28, 333-345 (2017)
10. Cueff, L., Huet, E., Schmitt, L., Pastezeur, S., Coquil, M., Savary, T., Pécréaux, J., Bouvrais, H.: Microtubule stiffening by doublecortin-domain protein ZYG-8 contributes to spindle orientation during C. elegans zygote division. bioRxiv 2024.2011.2029.624795 (2025)
11. Bicek, A.D., Tüzel, E., Kroll, D.M., Odde, D.J.: Analysis of Microtubule Curvature. Methods in Cell Biology, vol. 83, pp. 237-268. Academic Press (2007)
12. Risca, V.I., Wang, E.B., Chaudhuri, O., Chia, J.J., Geissler, P.L., Fletcher, D.A.: Actin filament curvature biases branching direction. PNAS 109, 2913-2918 (2012)
13. Pallavicini, C., Monastra, A., Bardeci, N.G., Wetzler, D., Levi, V., Bruno, L.: Characterization of microtubule buckling in living cells. Eur. Biophysics J. 46, 581-594 (2017)
14. Wisanpitayakorn, P., Mickolajczyk, K.J., Hancock, W.O., Vidali, L., Tüzel, E.: Measurement of the persistence length of cytoskeletal filaments using curvature distributions. Biophysical journal 121, 1813-1822 (2022)
15. Nishida, K., Matsumura, K., Tamura, M., Nakamichi, T., Shimamori, K., Kuragano, M., Kabir, A.M.R., Kakugo, A., Kotani, S., Nishishita, N., Tokuraku, K.: Effects of three microtubule-associated proteins (MAP2, MAP4, and Tau) on microtubules' physical properties and neurite morphology. Scientific Reports 13, 8870 (2023)
16. Ju, H., Skibbe, H., Fukui, M., Yoshimura, S.H., Naoki, H.: Machine learning-guided reconstruction of cytoskeleton network from live-cell AFM images. iScience 27, 110907 (2024)
17. Blob, A., Ventzke, D., Rölleke, U., Nies, G., Munk, A., Schaedel, L., Köster, S.: Global alignment and local curvature of microtubules in mouse fibroblasts are robust against perturbations of vimentin and actin. Soft matter (2025)
18. Pallavicini, C., Levi, V., Wetzler, D.E., Angiolini, J.F., Benseñor, L., Despósito, M.A., Bruno, L.: Lateral Motion and Bending of Microtubules Studied with a New Single-Filament Tracking Routine in Living Cells. Biophysical journal 106, 2625-2635 (2014)
19. Bouvrais, H., Crespo, M.: MicSim_FluoMT: Two synthetic datasets of images of fluorescent microtubules (Ait Laydi et al., 2025). Zenodo (2025)

20. Ait Laydi, A., Cueff, L., Crespo, M., Mourabit, Y.E., Bouvrais, H.: Adaptive Attention Residual U-Net for curvilinear structure segmentation in fluorescence microscopy and biomedical images. arXiv preprint arXiv:2507.07800 (2025)
21. Papereux, S., Leconte, L., Valades-Cruz, C.A., Liu, T., Dumont, J., Chen, Z., Salamero, J., Kervrann, C., Badoual, A.: DeepCristae, a CNN for the restoration of mitochondria cristae in live microscopy images. Communications Biology 8, 320 (2025)
22. Ebrahimi, V., Stephan, T., Kim, J., Carravilla, P., Eggeling, C., Jakobs, S., Han, K.Y.: Deep learning enables fast, gentle STED microscopy. Communications Biology 6, 674 (2023)
23. Chen, R., Tang, X., Zhao, Y., Shen, Z., Zhang, M., Shen, Y., Li, T., Chung, C.H.Y., Zhang, L., Wang, J., Cui, B., Fei, P., Guo, Y., Du, S., Yao, S.: Single-frame deep-learning super-resolution microscopy for intracellular dynamics imaging. Nature Comm. 14, 2854 (2023)
24. Mathieu, M., Couprie, C., LeCun, Y.: Deep multi-scale video prediction beyond mean square error. arXiv preprint arXiv:1511.05440 (2015)
25. Nie, D., Trullo, R., Lian, J., Wang, L., Petitjean, C., Ruan, S., Wang, Q., Shen, D.: Medical image synthesis with deep convolutional adversarial networks. IEEE Transactions on Biomedical Engineering 65, 2720-2730 (2018)
26. Siddique, N., Sidike, P., Elkin, C., Devabhaktuni, V.: U-Net and its variants for medical image segmentation: theory and applications. arXiv preprint arXiv:2011.01118 (2020)
27. Zhang, R., Jiang, G.: Exploring a multi-path U-net with probability distribution attention and cascade dilated convolution for precise retinal vessel segmentation in fundus images. Scientific Reports 15, 13428 (2025)
28. Wang, Z., Bovik, A.C.: Mean squared error: Love it or leave it? A new look at signal fidelity measures. IEEE signal processing magazine 26, 98-117 (2009)
29. Chen, H., He, X., Qing, L., Wu, Y., Ren, C., Sheriff, R.E., Zhu, C.: Real-world single image super-resolution: A brief review. Information Fusion 79, 124-145 (2022)
30. Botchkarev, A.: A new typology design of performance metrics to measure errors in machine learning regression algorithms. Interdisciplinary Journal of Information, Knowledge, and Management 14, 045-076 (2019)
31. von Chamier, L., *et al.* : Democratising deep learning for microscopy with ZeroCostDL4Mic. Nature Communications 12, 2276 (2021)
32. Lu, Z., Chen, Y.: Single image super-resolution based on a modified U-net with mixed gradient loss. signal, image and video processing 16, 1143-1151 (2022)
33. Ge, L., Dou, L.: G-Loss: A loss function with gradient information for super-resolution. Optik 280, 170750 (2023)
34. Chen, C., Yang, X., Huang, R., Hu, X., Huang, Y., Lu, X., Zhou, X., Luo, M., Ye, Y., Shuang, X., Miao, J., Xiong, Y., Ni, D.: Fine-Grained Correlation Loss for Regression. In: Medical Image Computing and Computer Assisted Intervention – MICCAI 2022, pp. 663-672. Springer Nature Switzerland, (2022)
35. Cueff, L., Pecreaux, J., Bouvrais, H.: MicReal_FluoMT: A dataset of microscopy images with stained microtubules (Ait Laydi et al., 2025). Zenodo (2025)
36. Staal, J., Abràmoff, M.D., Niemeijer, M., Viergever, M.A., Van Ginneken, B.: Ridge-based vessel segmentation in color images of the retina. IEEE transactions on medical imaging 23, 501-509 (2004)
37. iMed: CORN: Corneal nerve fiber dataset. Zenodo (2024)